\documentclass[11pt,a4paper]{article}
\usepackage[hyperref]{acl2020}
\usepackage{times}
\usepackage{latexsym}
\usepackage{amssymb}
\usepackage{amsmath}
\usepackage{multirow}
\usepackage{tablefootnote}
\usepackage{subcaption}
\usepackage{graphicx}
\usepackage{float}
\usepackage{framed}
\usepackage{paralist}
\usepackage{bm}
\usepackage{color,xcolor}
\usepackage{booktabs}

\usepackage{amsmath}
\usepackage{algorithm}
\usepackage[noend]{algorithmic}
\newcommand{\algorithmicinput}{\textbf{input}}
\newcommand{\INPUT}{\item[\algorithmicinput]}

\usepackage{microtype}

\aclfinalcopy % Uncomment this line for the final submission
\newcommand\Rouge{\textsc{Rouge}}

\title{Discrete Optimization for Unsupervised Sentence Summarization with Word-Level Extraction}

\author{Raphael Schumann$^1$, 
  Lili Mou$^2$,
  Yao Lu$^3$, 
  Olga Vechtomova$^3$,  
  Katja Markert$^1$ \\
  $^1$\normalsize Institute of Computational Linguistics, Heidelberg University, Germany \\
  \texttt{\normalsize \{rschuman, markert\}@cl.uni-heidelberg.de}\\ \normalsize
  $^2$Dept. Computing Science, University of Alberta, Canada; Alberta Machine Intelligence Institute (Amii)\\
  \texttt{\normalsize doublepower.mou@gmail.com} \\
  $^3$\normalsize University of Waterloo, Canada\\
  \texttt{\normalsize \{yao.lu, ovechtom\}@uwaterloo.ca}}

\date{}

\begin{document}
\maketitle
\begin{abstract}
Automatic sentence summarization produces a shorter version of a sentence, while preserving its most important information. A good summary is characterized by language fluency and high information overlap with the source sentence. We model these two aspects in an unsupervised objective function, consisting of language modeling and semantic similarity metrics. We search for a high-scoring summary by discrete optimization. Our proposed method achieves a new state-of-the art for unsupervised sentence summarization according to \Rouge\ scores. Additionally, we demonstrate that the commonly reported \Rouge\ F1 metric is sensitive to summary length. Since this is unwillingly exploited in recent work, we emphasize that future evaluation should explicitly group summarization systems by output length brackets.\footnote{Our code and system outputs are available at: \url{https://github.com/raphael-sch/HC_Sentence_Summarization}}
\end{abstract}

\setlength{\textfloatsep}{10pt}
\section{Introduction}
Sentence summarization 
transforms a long source sentence into a short summary, while preserving key information~\citep{rush2015neural}. Sentence summarization has wide applications, for example, news headline generation and text simplification.

State-of-the-art sentence summarization systems are based on sequence-to-sequence neural networks~\cite{rush2015neural, nallapati2016abstractive, wang-etal-2019-biset}, which 
require massive parallel data for training. 
Therefore, unsupervised sentence summarization has recently attracted increasing interest. Cycle-consistency approaches treat the summary as a discrete latent variable and use it to reconstruct the source sentence \citep{wang-lee-2018-learning, baziotis2019seq}. Such latent-space generation fails to explicitly model the resemblance between the source sentence and the target summary. \citet{zhou2019simple} propose a left-to-right beam search approach based on a heuristically defined scoring function. However, beam search is biased towards the first few words of the source. 
\begin{figure}[t]
    \centering
    \includegraphics[width=0.47\textwidth]{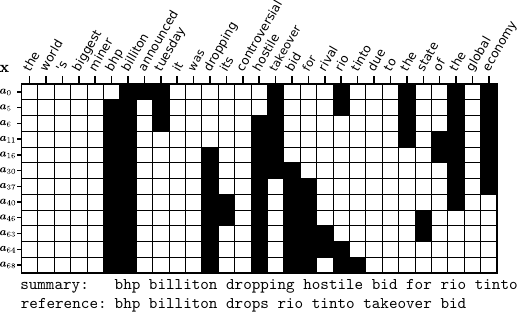}
    \caption{Summarizing a sentence $\mathbf{x}$ by hill climbing. Each row is a Boolean vector $\bm{a}_t$ at a search step $t$ . A black cell indicates a word is selected, and \textit{vice versa}. Randomly swapping two values in the Boolean vector yields a new summary that is scored by an objective function that measures language fluency and semantic similarity. If the new summary increases the objective, this summary is accepted as the current best solution. Rejected solutions are not depicted.}
    \label{fig:progress}
\end{figure}

In this paper, we propose a hill-climbing approach to unsupervised sentence summarization, directly extracting words from the source sentence. This is motivated by the observation that human-written reference summaries exhibit high word overlap with the source sentence, even preserving word order to a large extent. 
To perform word extraction for summarization, we define a scoring function --- similar to \newcite{miao2019cgmh} and \newcite{zhou2019simple} --- that evaluates the quality of a candidate summary by language fluency, semantic similarity to the source, and a hard constraint on output length. We search towards our scoring function by first choice hill-climbing (FCHC), shown in Figure~\ref{fig:progress}. We start from a random subset of words of the required output length. For each search step, a new candidate is sampled by randomly swapping a selected word and a non-selected word. We accept the new candidate if its score is higher than the current one. In contrast to beam search~\cite{zhou2019simple}, our summary is not generated sequentially from the beginning of a sentence, and therefore not biased towards the first few words.

Due to the nature of the search action, our approach is able to \textit{explicitly} control the length of a summary as a hard constraint. In all previous work, the summary length is weakly controlled by length embeddings or a soft length penalty \cite{zhou2019simple,wang-lee-2018-learning,fevry2018unsupervised,baziotis2019seq}.
Thus, the generated summaries by different systems vary considerably in average length, for example, ranging from 9 to 15 on a headline corpus (Section~\ref{sec:data}). Previous work uses \Rouge{} F1 to compare summaries that might differ in length. We show
that \Rouge{} F1 is unfortunately sensitive to summary output length, in general favoring models that produce longer summaries. Therefore, we argue that controlling the output length should be an integral part of the summarization task and that a fair system comparison can only be conducted between summaries in the same length bracket.

Our model establishes a new state-of-the-art for unsupervised sentence summarization across all commonly-used length brackets and different \Rouge{} metrics on the Gigaword dataset for headline generation \cite{rush2015neural} and on DUC2004 \cite{over2004introduction}.

The main contributions of this paper are:
\begin{compactitem}[$\bullet$]
\item We propose a novel method for unsupervised sentence summarization by hill climbing with word-level extraction. 
\item We outperform current unsupervised sentence summarization systems, including more complex sentence reconstruction models.
\item We show that \Rouge\ F1 is sensitive to summary length and thus emphasize the importance of explicitly controlling summary length for a fair comparison among different summarization systems.
\end{compactitem}
\section{Related Work}
\label{sec:related_work}

\textbf{Text Summarization.}
The task can be categorized by source text types, such as \textit{multi-document summarization}~\cite{erkan,radev-etal-2000-centroid,haghighi} and \textit{single-document summarization}~\cite{mihalcea-tarau-2004-textrank, zhou-hovy-2004-template, zheng-lapata-2019-sentence}.
Traditional approaches are mostly \textit{extractive}, i.e., they extract entire sentences from a document. Recently, sequence-to-sequence (Seq2Seq) models have been used for \textit{abstractive} summaries, where the system is able to synthesize new sentences \cite{nallapati2016abstractive, nallapati2017summarunner, gehrmann2018bottom, lewis2019bart,fabbri2019multi}. The copy mechanism~\cite{gu2016incorporating} in a Seq2Seq model can be viewed as word-level extraction in abstractive summarization \cite{see2017get, paulus2017deep}. Both state-of-the-art extractive and abstractive approaches are usually supervised.

\textit{Sentence summarization} yields a short summary for a long sentence. \citet{hori2004speech} and \citet{clarke2006constraint} extract single words from the source sentence based on language model fluency and linguistic constraints. They search via dynamic programming with a trigram language model, which restricts the model capacity.
The Hedge Trimmer method \cite{dorr2003hedge} also uses hand-crafted linguistic rules to remove constituents from a parse tree until a certain length is reached. 

\citet{rush2015neural} propose a supervised abstractive sentence summarization system with an attention mechanism \citep{bahdanau2014neural}, and they also introduce a dataset for headline generation derived from Gigaword.\footnote{\url{https://catalog.ldc.upenn.edu/LDC2003T05}} Subsequent models for this dataset were also supervised and mostly based on Seq2seq architectures \cite{nallapati2016abstractive, chopra2016abstractive, wang-etal-2019-biset}.

Recently, unsupervised approaches for sentence summarization have attracted increasing attention. \citet{fevry2018unsupervised} learn a denoising autoencoder and control the summary length by a length embedding. \citet{wang-lee-2018-learning} and \citet{baziotis2019seq} use cycle-consistency \cite{NIPS20166469} to learn the reconstruction of the source sentence and return the intermediate discrete representation as a summary. \citet{zhou2019simple} use beam search to optimize a scoring function, which considers language fluency and contextual matching.

Our work can be categorized under unsupervised sentence summarization. We accomplish this by word-level extraction from the source sentence.

\textbf{Constrained Sentence Generation.} 
Neural sentence generation is usually accomplished in an autoregressive way, for example, by recurrent neural networks generating words left-to-right. This is often enhanced by beam search \cite{sutskever2014sequence}, which keeps a beam of candidates in a partially greedy fashion.
A few studies allow hard constraints on this decoding procedure. \citet{hokamp2017lexically} use grid-beam search to impose lexical constraints during decoding. \citet{anderson-etal-2017-guided} propose constrained beam search to predict fixed image tags in an image transcription task. \citet{miao2019cgmh} propose a Metropolis--Hastings sampler for sentence generation, where hard constraints can be incorporated into the target distribution. This is further extended to simulated annealing~\cite{UPSA}, or applied to the text simplification task~\cite{simplification}. Different from the above concurrent work, this paper applies the stochastic search framework to text summarization, and design our specific search space and search actions for word extraction.

In previous work on text summarization, length embeddings~\cite{kikuchi-etal-2016-controlling,fan2018controllable} have been used to indicate the desired summary length. However, these are not hard constraints, because the model may learn to ignore such information.

\setlength\abovedisplayskip{0pt}
\setlength{\belowdisplayskip}{0pt}

\section{Proposed Model}
Given a source sentence $\mathbf{x}=(x_1, x_2, \dots, x_n)$ as input, our goal is to generate a shorter sentence $\mathbf{y}=(y_1, y_2, \dots, y_m)$ as a summary of $\mathbf{x}$. We perform word-level extraction, in addition keeping the original word order intact. Thus, $\mathbf{y}$ is a subsequence of $\mathbf x$. Our word-level extraction optimizes a manually defined objective function $f(\mathbf{y};\mathbf{x}, s)$, where the summary length $s$ is predefined ($s<n$) and not subject to optimization. In the remainder of this section, we will describe the objective function, search space, and the search algorithm in detail.
\subsection{Search Objective}

We define an objective function $f(\mathbf y;\mathbf x, s)$, which our algorithm maximizes. It evaluates the fitness of a candidate sentence $\mathbf{y}$ as the summary of an input $\mathbf{x}$, involving three aspects, namely, language fluency $f_{\overleftrightarrow{\mathrm{LM}}}(\mathbf{y})$, semantic similarity $f_{\mathrm{SIM}}(\mathbf{y};\mathbf{x})$, and a length constraint $f_{\mathrm{LEN}}(\mathbf{y}, s)$. This is given by

\small
\begin{equation}
f(\mathbf y;\mathbf x,s)=f_{\overleftrightarrow{\mathrm{LM}}}(\mathbf{y})\cdot f_{\mathrm{SIM}}(\mathbf{y};\mathbf{x})^\gamma \cdot f_{\mathrm{LEN}}(\mathbf{y}; s),
\end{equation} 

\vspace{.2cm}
\normalsize\noindent where the relative weight $\gamma$ balances $f_{\overleftrightarrow{\mathrm{LM}}}(\mathbf{y})$ and $f_{\mathrm{SIM}}(\mathbf{y};\mathbf{x})$. We treat the summary length as a hard constraint, and therefore we do not need a weighting hyperparameter for $f_{\mathrm{LEN}}$.

\textbf{Language Fluency.}
The language fluency scorer quantifies how grammatical and idiomatic a candidate summary $\mathbf{y}$ is. Our model generates a candidate summary in a non-autoregressive fashion, in contrast to the beam search in \citet{zhou2019simple}. Thus, we are able to simultaneously consider forward and backward language models, using the geometric average of their perplexities. Using both forward and backward language models is less biased towards sentence beginnings or endings.\vspace{-.3cm}

\small
\begin{equation}\nonumber
\overleftrightarrow{\mathrm{PPL}}(\mathbf{y})=\sqrt[\leftroot{-2}\uproot{2}2|\mathbf{y}|]{\prod_{i}^{|\mathbf{y}|} \frac{1}{p_{\overrightarrow{\mathrm{LM}}}(y_i|\mathbf{y}_{<i})}\prod_{i}^{|\mathbf{y}|} \frac{1}{p_{\overleftarrow{\mathrm{LM}}}(y_i|\mathbf{y}_{>i})}}.
\end{equation}
\normalsize Our fluency scorer is the inverse perplexity.

\small\begin{equation}
f_{\overleftrightarrow{\mathrm{LM}}}(\mathbf{y})={\overleftrightarrow{\mathrm{PPL}}(\mathbf{y})}^{-1}.
\end{equation}

\normalsize Depending on applications, the language models could be pretrained on a target corpus.\footnote{We use the terminology \textit{unsupervised summarization}, following \citet{zhou2019simple}. While we train the language models on the desired target language, we do not need parallel source-target pairs, i.e., sentences together with their groundtruth summaries.} In this case, the fluency scorer also measures 
whether the summary style is consistent with the target language.
This could be important in certain applications, e.g., headline generation, where the summary language differs from the input in style.

\textbf{Semantic Similarity.}
A semantic similarity scorer ensures that the summary keeps the key information of the input sentence. We adopt the cosine similarity between sentence embeddings as\vspace{-.2cm}

\small\begin{equation}
f_{\mathrm{SIM}}(\mathbf{y}; \mathbf{x})=\cos(\bm e(\mathbf{x}), \bm e(\mathbf{y})),
\end{equation}
\normalsize where $\bm e$ is a sentence embedding method. In our work, we use unigram word embeddings learned by the sent2vec model \cite{pagliardini-etal-2018-unsupervised}.
Then, $\bm{e}(\mathbf x)$ is computed as the average of these unigram embeddings, weighted by the inverse-document frequency (\textit{idf}) of the words.

We use sent2vec because it is trained in an unsupervised way on individual sentences. By contrast, other unsupervised methods like SiameseCBOW \cite{kenter-etal-2016-siamese} or BERT \cite{devlin-etal-2019-bert} use adjacent sentences as part of the training signal.

\textbf{Length Constraint.} \label{sss:length}
Our discrete searching approach is able to impose the output length as a hard constraint, allowing the model to generate summaries of any given length.
Suppose the desired output length is $s$, then our length scorer is\vspace{-.2cm}

\small\begin{equation}
f_{\mathrm{LEN}}(\mathbf{y}; s)=
\begin{cases}
  1, & \text{if}\ |\textbf{y}|=s, \\
  -\infty, & \text{otherwise}.
\end{cases}
\end{equation}
\normalsize In other words, a candidate summary $\mathbf y$ is \textit{infeasible} if it does not satisfy the length constraint. In practice, we implement this hard constraint by searching among \textit{feasible solutions} only.

\subsection{Search Space}

Most sentence generation models choose a word from the vocabulary at each time step, such as autoregressive generation that predicts the next word~\cite{sutskever2014sequence,rush2015neural}, and edit-based generation with deletion or insertion operations~\cite{miao2019cgmh,dong}. In these cases, the search space is $|\mathcal{V}|^s$, given a vocabulary $\mathcal{V}$ and a summary length $s$.

However, reference summaries are highly extractive. In the headline generation dataset~\cite{rush2015neural}, for example, 45\% of the words in the reference summary also appear in the source sentence. This yields a ceiling of 45 \mbox{\Rouge{}-1} F1 points\footnote{We assume an extracted summary has the same length as the reference, and 45\% words of the reference are in the original sentence. This gives us a ceiling of 45\% precision and recall.} for a purely extractive method, which is higher than the current state-of-the-art supervised abstractive result of 39~points~\cite{wang-etal-2019-biset}. 
We are thus motivated to propose our word-extraction approach that extracts a subsequence of the input as the summary.
 Additionally, we arrange the words in the same order as the input, motivated by the \textit{monotonicity assumption} in summarization~\citep{yu2016online, raffel2017online}.

Formally, we define the search space as $\bm a = (a_1, \dots, a_n) \in\{0,1\}^n$, where $n$ is the length of the input sentence $\mathbf x$. The vector $\bm a$ is a Boolean filter over the source words $\mathbf{x}$. The summary sequence can then be represented by $\mathbf{y}=\mathbf{x}_{\bm a}$, i.e., we sequentially extract words from the source sequence $\mathbf {x}$ by the Boolean vector $\bm a$. If $a_i=1$, then $x_i$ is extracted for the summary, and {vice versa}. 

Further, we only consider the search space of all feasible solutions $\{\bm a:f(\mathbf x_{\bm a};\mathbf x, s) > -\infty$\}. That is to say, the candidate summary has to satisfy the length constraint in Section~\ref{sss:length}. Equivalently, the output length can be expressed by a constraint on the search space such that $\sum_i a_i=s$.

The above restrictions reduce the search space to $\binom{n}{s}$ solutions. In a realistic setting, our search space is much smaller than that of generating words from the entire vocabulary.

\small\begin{algorithm}[t] \small
    \caption{First-Choice Hill Climbing}
    \label{algorithm:shc}
    \begin{algorithmic}
    \INPUT{objective~function~$f(\mathbf{y};\mathbf{x}, s)$, source~sentence~$\mathbf{x}$, summary~length~$s$, number~of~steps~$T$, initial~random~solution~$\bm{a}_0$, neighbor~function~$q(\bm a'|\bm a)$}
    \FOR{$t = 1$ to $T$}
        \STATE $\mathbf{y}_{t-1}=\mathbf{x}_{\bm a_{t-1}}$
        \STATE $\bm a' \sim q(\cdot|\bm a_{t-1})$
        \STATE $\mathbf{y}'  =  \mathbf{x}_{\bm a'}$
        \IF{${f(\mathbf{y}';\mathbf{x}, s)}\ge{f(\mathbf{y}_{t-1};\mathbf x,s)}$}
        \STATE{$\bm a_t =\bm a'$}
        \ELSE 
        \STATE{$\bm a_t =\bm a_{t-1}$}
        \ENDIF
    \ENDFOR
    \RETURN $\mathbf{y}^*\xleftarrow{}\mathbf x_{\bm a_T}$
\end{algorithmic}
\end{algorithm}
\normalsize\subsection{Search Algorithm}
We optimize our objective function $f(\mathbf{y};\mathbf{x}, s)$ by first-choice hill climbing \cite[FCHC,][]{russell2016artificial}. This is a stochastic optimization algorithm that proposes a candidate solution by local change at every search step. The candidate is accepted if it is better than the current solution. Otherwise, the algorithm keeps the current solution. FCHC maximizes the objective function in a greedy fashion and yields a (possibly local) optimum.

Algorithm~\ref{algorithm:shc} shows the optimization procedure of our FCHC. For each search step, a new candidate is sampled from the 
neighbor function $q(\bm a'|\bm a)$.
This is accomplished by randomly swapping two actions $a_i$ and $a_j$ for $a_i\ne a_j$, i.e., replacing a word in the summary with a word from the source sentence that is not in the current summary. The order of selected words is kept
as in the source sentence.
If the candidate solution achieves a higher score, then it is accepted. Otherwise, the candidate is rejected and the algorithm proceeds with the current solution. Our search terminates if it exceeds a pre-defined budget. The last solution is returned as the summary, as it is also the best-scored candidate due to our greedy algorithm.

One main potential drawback of hill climbing algorithms is that they may get stuck in a local optimum. To alleviate this problem, we restart the algorithm with multiple random initial word selections $\bm a_0$ and return the overall best solution. We set the number of restarts as \mbox{$\beta_R\cdot ns^2$} and number of search steps as \mbox{$\beta_T\cdot ns^2$}, where $\beta_R$ and $\beta_T$ are controlling hyperparameters. We design the formula to encourage more search for longer input sentences, but only with a tractable growth: linear for input length and quadratic for summary length. As the summary length is usually much smaller than the input length, quadratic search is possible. Increasing the number of restarts (and search steps) monotonically improves the scoring function, and thus in practice can be set according to the available search budget.

Other discrete optimization algorithms can be explored for sentence generation, such as simulated annealing~\cite{UPSA} and genetic algorithms. Our analysis on short sentences (where exhaustive search is tractable) showed that hill climbing with restarts achieves \Rouge{} scores similar to exhaustive search (Section~\ref{sec:analysis}).

\section{Evaluation Framework}
\label{sec:evaluation}
In this section, we will describe the datasets, evaluation metrics, and a widely used baseline (called Lead). 
Additionally, we report the observation that the commonly used evaluation metric, \Rouge{} F1, is sensitive to summary length, preferring longer summaries. Thus, we propose to group models with similar output length during evaluation for fair comparison.

\subsection{Datasets}\label{sec:data}
We evaluate our models on the dataset provided for DUC2004 Task 1 \citep{over2004introduction} and a headline generation corpus\footnote{\url{https://github.com/harvardnlp/NAMAS}} \cite{rush2015neural}, both widely adopted in the summarization literature. 

The DUC2004 dataset is designed and used for testing only. It consists of 500 news articles, each paired with four human written summaries. We follow \citet{rush2015neural} and adopt DUC2004 for sentence summarization by using only the first sentence of an article as input. The reference summaries are around 10 words long on average.

The headline generation dataset \cite{rush2015neural} is derived from the Gigaword news corpus. Each headline/title is viewed as the reference summary of the first sentence of an article. The dataset contains 3.8M training instances and 1951 test instances. The average headline contains $\sim$8 words; the average source sentence contains $\sim$30 words. We use 500 held-out validation instances for hyperparameter tuning. Note that the training set is only used to train a language model and sent2vec embeddings. The summarization process itself is not trained in our approach.

\subsection{Lead Baselines}\label{ss:lead}
Lead baselines are a strong competitor that extracts the first few characters or words of the input sentence.
The DUC2004 shared task includes a Lead baseline, which extracts the first 75 characters as the summary. We call it Lead-C-75. For the Gigaword dataset, the reference has 8 words on average, and it is common to compare with a Lead variant that chooses the first 8 words. We call this baseline Lead-N-$n$ when we choose $n$ words. 
For fair comparison with previous work \cite{baziotis2019seq,fevry2018unsupervised} in Section~\ref{sec:competing_models}, we further introduce a new variant that returns the first $p$ percent of source words as the summary. We denote this baseline by {{Lead-P}-$p$}.

\subsection{\Rouge{} Scores}

Summarization systems are commonly evaluated by \Rouge{} scores \citep{lin2004rouge}. The \Rouge{}-1 (or \Rouge{}-2) score computes the unigram (or bigram) overlap of a generated summary and the reference. \Rouge{}-L calculates the longest common subsequence. Depending on the dataset, either \Rouge{} Recall or \Rouge{} F1 variant is adopted. Since the \Rouge{} Recall metric is not normalized with regard to length, DUC2004 standard evaluation truncates the summary at 75 characters. This procedure was also adopted by \citet{rush2015neural} for the headline generation task, but later \citet{chopra2016abstractive} proposed to report the ``more balanced" \Rouge{} F1 metric for the Gigaword headline generation dataset and abandoned truncation. We follow previous work and use {{\Rouge{} F1} for headline generation and \mbox{{truncated \Rouge{} Recall}} for DUC2004}.

\subsection{Summary Length}\label{section:sumary_length}
\begin{figure}[!t]
  \centering
     \begin{subfigure}{0.49\textwidth}
  \includegraphics[width=\textwidth]{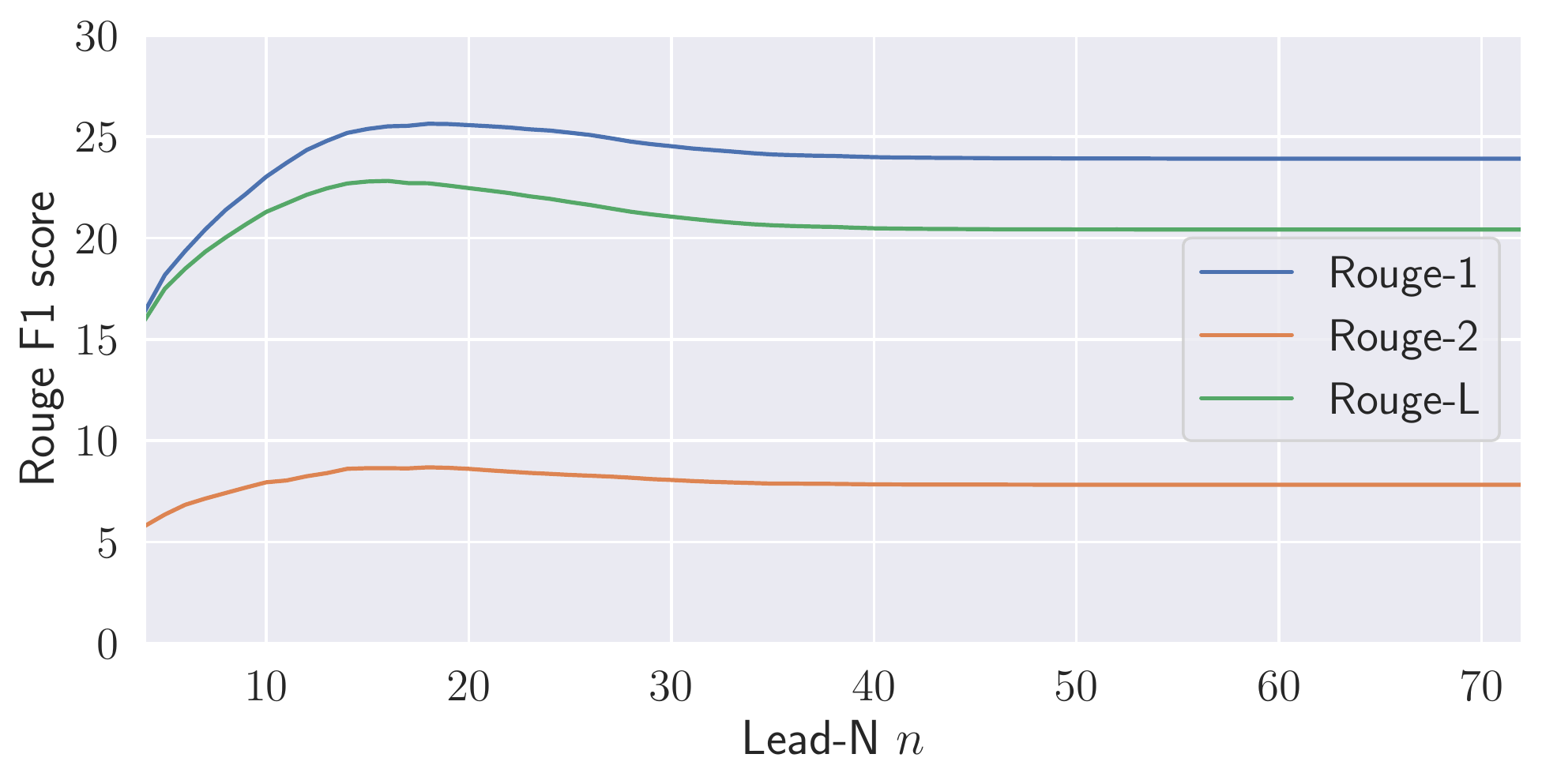}
  \label{fig:lead_n_f1}
  \end{subfigure}
  \hfill
  \vspace*{-0.7cm}
  \begin{subfigure}{0.49\textwidth}
  \includegraphics[width=\textwidth]{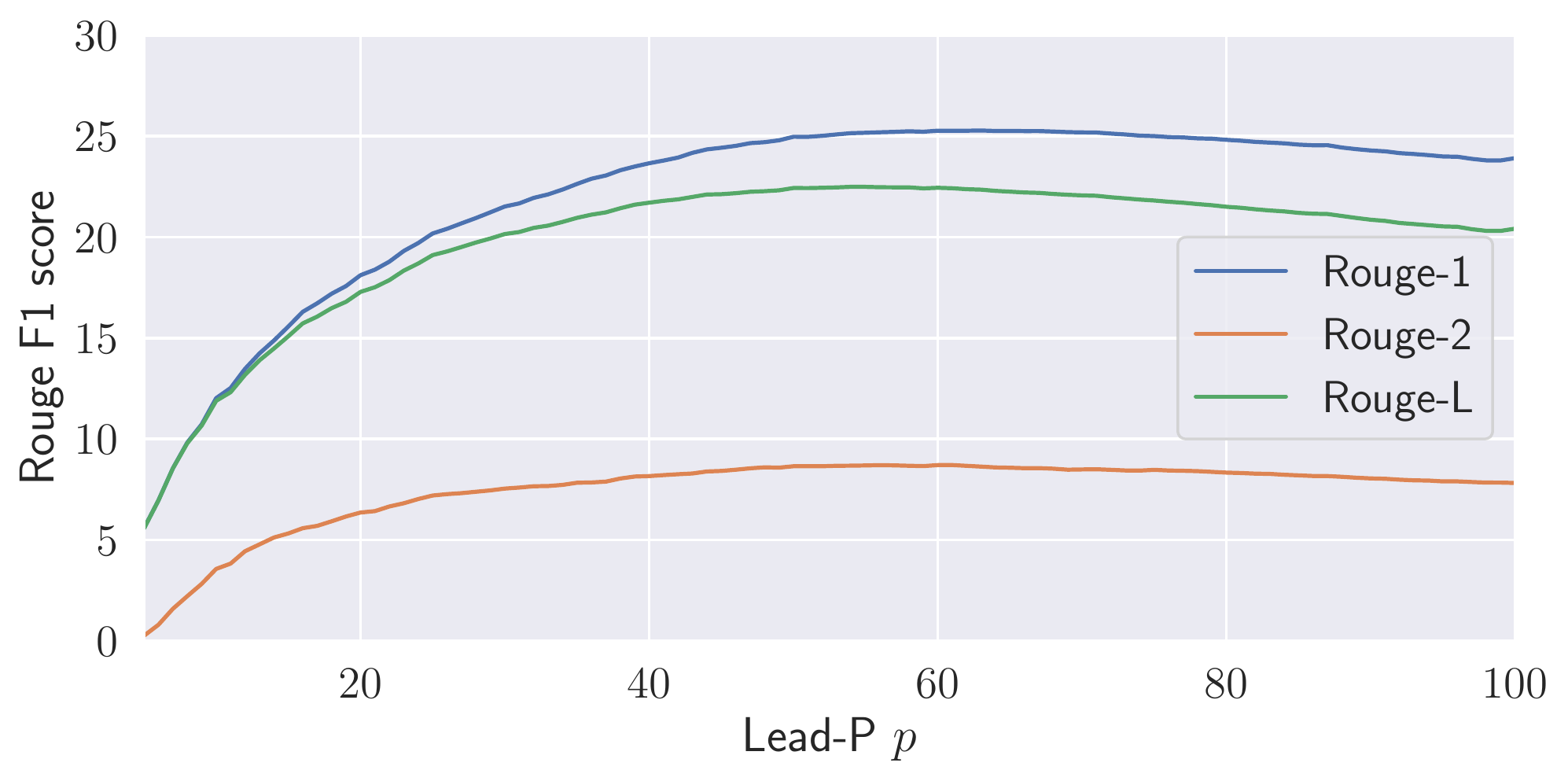}
  \label{fig:lead_p_f1}
  \end{subfigure}
\vspace{-0.7cm}
\caption{\Rouge{} F1 scores on the test set of headline generation for Lead-N and Lead-P baselines with different number $n$ and percentage $p$ of leading words.}
\label{figure:leadn_leadp}
\end{figure}

As mentioned, \Rouge{} F1 was introduced to the evaluation of sentence summarization to better compare models with different output lengths~\cite{chopra2016abstractive, nallapati2016abstractive}. To investigate the effect of summary length on \Rouge{} F1, we calculate \Rouge{} F1 scores for the Lead-N-$n$ and Lead-P-$p$ baselines with different length parameters. Figure~\ref{figure:leadn_leadp} shows that \Rouge{} F1 peaks at $n\approx18$ or $p\approx50$. The difference between the maximum performance at $n\approx18$ and the widely adopted baseline (Lead-N-$8$) is large: 4.2 \Rouge-1 F1 points. 
A similar effect is observed by \citet{sun-etal-2019-compare} for document summarization. This shows that \Rouge{} F1 is still sensitive to summary length, and this effect should be considered during evaluation. We propose to report the average output length of a model and only compare models in the same length bracket.

\section{Experiments}

\subsection{Setup}

We conduct experiments with two settings, dependent on how the scorers $f_{\overleftrightarrow{\mathrm{LM}}}$ and $f_{\mathrm{SIM}}$ are trained. 
In the first setting, we train the language model and sent2vec embeddings on the \textbf{source} (article) side of the Gigaword headline generation dataset. This complies with \newcite{fevry2018unsupervised} and \newcite{baziotis2019seq}.
In the second setting, we train the language model and sent2vec embeddings on the \textbf{target} (title) side like \citet{zhou2019simple}. In both settings, we do not need parallel source-target pairs.

For output length, our headline generation experiment sets the desired target length as 8 words, 10 words, and 50\% of the input, as these mirror
either the average reference summary length or the average output lengths of our competitors~\cite{wang-lee-2018-learning,zhou2019simple,fevry2018unsupervised,baziotis2019seq}.
For DUC2004, the desired summary length is set to 13 words, because the standard evaluation script truncates after the first 75 characters (roughly 13 words) in the summary.

Our forward and backward language models use long short term memory units \citep{hochreiter1997long} and are optimized for 50 epochs by stochastic gradient descent. Embeddings and hidden sizes are set to 1024 dimensions.

We tune hyperparameters on the development data of the headline corpus, and set the weighting parameter $\gamma$ to 12 for all models. The search steps and restarts are set to \mbox{$\beta_T=0.1$} and \mbox{$\beta_R=0.035$}, respectively. We see a sharp performance improvement when we do more searching. Thus, we choose $\beta_T$ and $\beta_R$ at the critical values due to efficiency concerns.

\begin{table*}
\centering

\resizebox{.85\textwidth}{!}{
\begin{tabular}{cl|ccc|r|rrr|r}
\hline
\multicolumn{2}{c|}{\textbf{Model}}             & \multicolumn{3}{c|}{\textbf{Data}}   & \multicolumn{1}{c|}{\textbf{Len D}} & \multicolumn{3}{c|}{\textbf{\Rouge{} F1}}                                                           & \multicolumn{1}{c}{\textbf{Len O}} \\ \hline
\multicolumn{2}{l|}{}                           & \textbf{article} & \textbf{title} & \textbf{external} & \multicolumn{1}{l|}{\textbf{}}      & \multicolumn{1}{c}{\textbf{R-1}} & \multicolumn{1}{c}{\textbf{R-2}} & \multicolumn{1}{c|}{\textbf{R-L}}                 & \multicolumn{1}{l}{\textbf{}}      \\ \hline
\multirow{3}{*}{A} & Lead-N-8                 & \checkmark          &            &            & 8                                   & 21.39                            & 7.42                             & 20.03                                                                         & 7.9                                \\
                     & \textit{HC\_article\_8}                     & \checkmark           &            &            & 8                                   & \underline{23.09}                                 & \underline{7.50}                                 & \underline{21.29}                                                                                 & 7.9                                   \\
                     & \textit{HC\_title\_8}                     &            & \checkmark           &            & 8                                   & \textbf{26.32}                                 & \textbf{9.63}                                 & \textbf{24.19}                                                                                 & 7.9                                   \\ \hline
\multirow{7}{*}{B}& Lead-N-10     & \checkmark           &           &            & 10                                   & 23.03                            & 7.95                             & 21.29                                                                         & 9.8                               \\ 
                    & \citet{wang-lee-2018-learning}     & \checkmark           & \checkmark           &            & -                                   & 27.29                            & 10.01                             & 24.59                                                                         & 10.8                               \\
                     & \citet{zhou2019simple}     &            & \checkmark           & billion    & -                                   & 26.48                            & 10.05                            & 24.41                                                                        & 9.3                                \\
                     & \textit{HC\_article\_10}                      & \checkmark           &            &            & 10                                  & 24.44                                 & 8.01                                 & 22.21                                                                                 & 9.8                                   \\
                     & \textit{HC\_title\_10}                      &            & \checkmark           &            & 10                                  & 27.52                            & 10.27                            & 24.91                                                                        & 9.8                                \\
                     & \textit{HC\_title+twitter\_10}                     &            & \checkmark           & twitter    & 10                                  & \underline{28.26}                                 & \underline{10.42}                                 & \underline{25.43}                                                                                 &    9.8                                \\
                     & \textit{HC\_title+billion\_10}                     &            &\checkmark          & billion    & 10                                  & \textbf{28.80}                                 & \textbf{10.66}                                 & \textbf{25.82}                                                                                 &    9.8                                \\\hline
\multirow{5}{*}{C} & Lead-P-50                & \checkmark           &            &            & 50\%                                & 24.97                            & \underline{8.65}                             & 22.43                                                                         & 14.6                               \\
                     & \citet{fevry2018unsupervised}  & \checkmark           &            & SNLI        & 50\%                                & 23.16                            & 5.93                             & 20.11                                                                        & 14.8                               \\
                     & \citet{baziotis2019seq} & \checkmark          &            &            & 50\%                                & 24.70                            & 7.97                             & 22.14                                                                         & 15.1                               \\
                     & \textit{HC\_article\_50p}                     & \checkmark          &            &            & 50\%                                & \underline{25.58}                                 & 8.44                                 & \underline{22.66}                                                                                 & 14.9                                   \\ 
                     & \textit{HC\_title\_50p}                     &           &    \checkmark        &            & 50\%                                & \textbf{27.05}                                 & \textbf{9.75}                                 & \textbf{23.89}                                                                                 & 14.9                                   \\\hline
\end{tabular}
}\vspace{-.2cm}
\caption{Results for headline generation on the Gigaword test set. \textbf{Data}: data used during training (source article, target titles, external corpus). \textit{billion}: the Billion Word Corpus~\citep{chelba2013one}; \textit{twitter}: the Twitter corpus~\citep{pagliardini-etal-2018-unsupervised}; \textit{SNLI}: the Stanford Natural Language Inference dataset~\citep{snli:emnlp2015}.
\textbf{Len~D}: desired summary length. \textbf{\Rouge\ F1~(R-1,~R-2,~R-L)}: \Rouge-1, \Rouge-2, \Rouge-L F1 scores. \textbf{Len~O}: averaged output length. \textbf{Best} results in bold. \underline{Second best} results underlined. \textbf{A}: Models with output length around 8 words. \textbf{B}: Models with output length around 10 words. \textbf{C}: Models with output length around 50\% of the input. Our hill-climbing (HC) approaches are named in the format of \textit{HC\_data\_outputLength}. }
\label{table:ext_giga}\vspace{-.3cm}
\end{table*}

\begin{table}[!t]
\centering\small

\resizebox{\linewidth}{!}{
\begin{tabular}{l|rrr}
\hline
\multicolumn{1}{c|}{\textbf{Model}} & \multicolumn{3}{c}{\textbf{\Rouge{} Recall}}                                                       \\ \hline
\textbf{}                             & \multicolumn{1}{c}{\textbf{R-1}} & \multicolumn{1}{c}{\textbf{R-2}} & \multicolumn{1}{c}{\textbf{R-L}} \\ \hline
    Lead-C-75                                            & 22.50                            & 6.49                             & 19.72                            \\
SEQ3  \cite{baziotis2019seq}                                     & 22.13                            & 6.18                             & 19.3                            \\
\textsc{Topiary} \cite{topiary}                   & 25.12                            & 6.46                             & 20.12                            \\
\textsc{BottleSum Ex} \cite{west-etal-2019-bottlesum}                              & 22.85                            & 5.71                             & 19.87                            \\
\textit{HC\_article\_13}                                                  & 24.21                                 & 6.63                                 & 21.24                                 \\
\textit{HC\_title\_13}                                                & \underline{26.04}                                 & \underline{8.06}                                 & \underline{22.90}                                 \\
\textit{HC\_title+twitter\_13}                                                & \textbf{27.41}                                 & \textbf{8.76}                                 & \textbf{23.89}                                 \\\hline
\end{tabular}\vspace{-.3cm}
}
\caption{Results on the DUC2004 dataset.}
\label{table:ext_DUC2004}\vspace{.1cm}
\end{table}
 
\subsection{Competing Models}
\label{sec:competing_models}
Besides the {Lead} baselines discussed in Section~\ref{ss:lead}, we compare our models with state-of-the-art unsupervised sentence summarization systems.

\citet{wang-lee-2018-learning}\footnote{Generated summaries are obtained via E-Mail correspondence. Scores differ because of evaluation setup.} use cycle-consistency to reconstruct source sentences from the headline generation corpus \cite{rush2015neural}. The latent discrete representation, learned to be similar to (non-parallel) headlines, is used as the summary.

\citet{zhou2019simple} optimize an objective function involving language fluency and contextual matching. Their language modeling scorer is trained on headlines of the Gigaword training set; their contextual matching scorer is based on ELMo embeddings \cite{peters-etal-2018-deep} trained with the Billion Word corpus \citep{chelba2013one}. Their summary length is controlled by a soft length penalty during beam search.

\citet{fevry2018unsupervised}\footnote{Retrained with official code (\url{https://github.com/zphang/usc_dae}) because the authors use a private test set.} learn a denoising autoencoder \cite{vincent2008extracting} to reconstruct source sentences of the Gigaword training set. Summary length is set to 50\% of the input length and is controlled by length embeddings in the decoder.

\citet{baziotis2019seq}\footnote{Retrained with official code (\url{https://github.com/cbaziotis/seq3}), because of different test data. The authors remove 54 noisy instances. Our replication thus achieves slightly lower scores than theirs.} propose \texttt{SEQ$^3$} that uses cycle-consistency to reconstruct source sentences from the Gigaword training set. The length is also set to 50\% of the input length, controlled by length embeddings in the intermediate decoder.

For the DUC2004 dataset, \textsc{Topiary} \cite{topiary} is the winning system in the competition. They shorten the sentence by rule-based syntax-tree trimming \cite{dorr2003hedge}, but enhance the resulting summary with topics that are learned on full articles.

\textsc{BottleSum Ex} \cite{west-etal-2019-bottlesum} uses the information bottleneck principle to predict the next sentence in an article. Their method employs a pretrained {small} GPT-2 model \cite{radford2019language}.

\subsection{Results}

\textbf{Results for Headline Generation.}
We first compare with Lead-N-8 (Group A, Table~\ref{table:ext_giga}). This is a standard baseline in previous work, because the average reference summary contains eight words. Unfortunately, none of the previous papers consider output length during evaluation, making comparisons between their (longer) output summaries and the Lead-N-8 baseline unfair, as discussed in Section~\ref{section:sumary_length}. Our approach, which explicitly controls summary length, considerably outperforms the Lead-N-8 baseline in a fair setting. 

Next, we compare with state-of-the-art unsupervised methods, whose output summary has roughly 10 words on average (Group B). In this case, we set our hard length constraint as 10 and include the Lead-N-10 baseline for comparison. 
Trained on the title side only, our \textit{HC\_title\_10} model outperforms these competing methods in all \Rouge{} F1 scores. In particular, \newcite{zhou2019simple} use the target side to train the language model, plus the Billion Word Corpus to pretrain embeddings used in the contextual matching scorer. With the same extra corpus to pretrain our sent2vec embeddings, our \textit{HC\_title+billion\_10} variant achieves even better performance, outperforming \citet{zhou2019simple} by 2.32 \Rouge-1 and 1.41 \Rouge-L points.

The Billion Word Corpus, however, includes complete articles, which implicitly yields unaligned parallel data. This could be inappropriate for an unsupervised method. Thus, we further train sent2vec embeddings on the Twitter corpus by \citet{pagliardini-etal-2018-unsupervised}.
The \textit{HC\_title+twitter\_10} also performs better than \textit{HC\_title\_10} and other competitors.

In Group C, we compare with the models whose summaries have an average length of 50\% of the input sentence. We set our desired target length to 50\% as well, and include the Lead-P-50 baseline. Previous studies report a performance improvement over the {Lead-N-8} baseline, but in fact, Table~\ref{table:ext_giga} shows that they do not outperform the appropriate Lead baseline Lead-P-50. Our model is the only unsupervised summarization system that outperforms the Lead-P-50 baseline on this dataset, even though it is trained solely on the article side.

It is noted that our models trained on the title side (\textit{HC\_title}) consistently outperform those trained on the article side (\textit{HC\_article}). This is not surprising because the former can generate headlines from the learned target distribution. This shows the importance of learning a summary language model even if we do not have supervision of parallel source-target data.

\textbf{Results for DUC2004.}
Table~\ref{table:ext_DUC2004} shows the results on the DUC2004 data. As this dataset is for test only, we directly transfer the models \textit{HC\_article} and \textit{HC\_title} from the headline generation corpus with the same hyperparameters (except for length). As shown in the table, we outperform all previous methods and the Lead-C-75 baseline. The results are consistent with Table~\ref{table:ext_giga}, showing the generalizability of our approach.

\textbf{Human Evaluation.} We conduct human evaluation via pairwise comparison of system outputs, in the same vein as \cite{west-etal-2019-bottlesum}. The annotator sees the source sentence along with the headline generated by our system and a competing method, presented in random order. The annotator is asked to compare the fidelity and fluency of the two systems, choosing among the three options (i) the first headline is better (ii) the second headline is better, and (iii) both headlines are equally good/bad. This task is repeated for 100 instances with 5 annotators each. The final label is selected by majority voting. The inter-annotator agreement (Krippendorff's alpha) is 0.25 when our model is compared with \newcite{wang-lee-2018-learning} and 0.17 with~\newcite{zhou2019simple}.

We report the aggregated score of our system in Table~\ref{fig:human}. For each sample, we count 1 point if our model wins, 0 points if it ties, -1 point if it loses. The points are normalized by the number of samples. The results show an advantage of our model over \newcite{wang-lee-2018-learning}, especially in fluency. Our model is also on par with \citet{zhou2019simple}. Note again that we achieve this with fewer data.

\begin{table}
\centering\small
\resizebox{\linewidth}{!}{
\begin{tabular}{l|lr}
\hline
\multicolumn{1}{c|}{\textbf{Models}} & \multicolumn{2}{c}{\textbf{Score (\#wins/\#ties/\#loses)}}                       \\ \hline
         & \multicolumn{1}{c|}{\textbf{Fidelity}} & \multicolumn{1}{c}{\textbf{Fluency}} \\ \hline
HC vs.~WL                 & \multicolumn{1}{l|}{+0.18 (44/30/26)}   & +0.30 (45/40/15)                      \\
HC vs.~ZR               & \multicolumn{1}{l|}{+0.05 (35/35/30)}   & -0.03 (24/49/27)       \\
\hline
\end{tabular}
}
\caption{Human evaluation in a pairwise comparison setting on 100 headline generation instances. We show the scores of our model (\textit{HC\_title\_10}) when it is compared with WL \cite{wang-lee-2018-learning} and ZR \cite{zhou2019simple}, in terms of average score of fidelity and fluency: 1 (wins), 0 (ties), and -1 (loses).}
\label{fig:human}
\end{table}
\subsection{Analysis}\label{sec:analysis}
In this section, we conduct  an in-depth analysis of our model, based on \textit{HC\_title\_10} for headline generation.

\begin{table}
\centering
\resizebox{.85\linewidth}{!}{
\begin{tabular}{l|rrr}
\hline
\multicolumn{1}{c|}{\textbf{Objective}}     & \multicolumn{3}{c}{\textbf{\Rouge\ F1 scores}}                                                             \\ \hline
\textbf{$f=$}                           & \multicolumn{1}{c}{\textbf{R-1}} & \multicolumn{1}{c}{\textbf{R-2}} & \multicolumn{1}{c}{\textbf{R-L}}                  \\ \hline
$f_{\overleftrightarrow{\mathrm{LM}}}  \cdot f_{\mathrm{SIM}}$ (full model)& \textbf{27.52}                            & \textbf{10.27}                            & \textbf{24.91}                                                      \\
$f_{\overrightarrow{\mathrm{LM}}} \cdot f_{\mathrm{SIM}}$     & 27.50                            & 10.15                            & 24.79                                                      \\
$f_{\overleftrightarrow{\mathrm{LM}}}$           & 25.24                            & 8.87                             & 23.09                                         \\
$f_{\overrightarrow{\mathrm{LM}}}$               & 25.18                            & 8.72                             & 22.93                                         \\
$f_{\mathrm{SIM}}$                               & 20.31                            & 4.08                             & 18.19                                         \\ \hline
\end{tabular}}
\caption{Ablation study of the search objective. Model \textit{HC\_title\_10} on the headline generation test set. Length constraint term omitted from notation.}
\label{table:ablation}
\end{table}

\textbf{Search Objective.} Table~\ref{table:ablation} provides an ablation study on our objective function. It shows that both language fluency and semantic similarity play a role in measuring the quality of a summary. The bi-directional language model is also slightly better than a uni-directional language model. 

\begin{figure}[t]
    \centering
    \vspace{-.2cm}
    \includegraphics[width=0.49\textwidth]{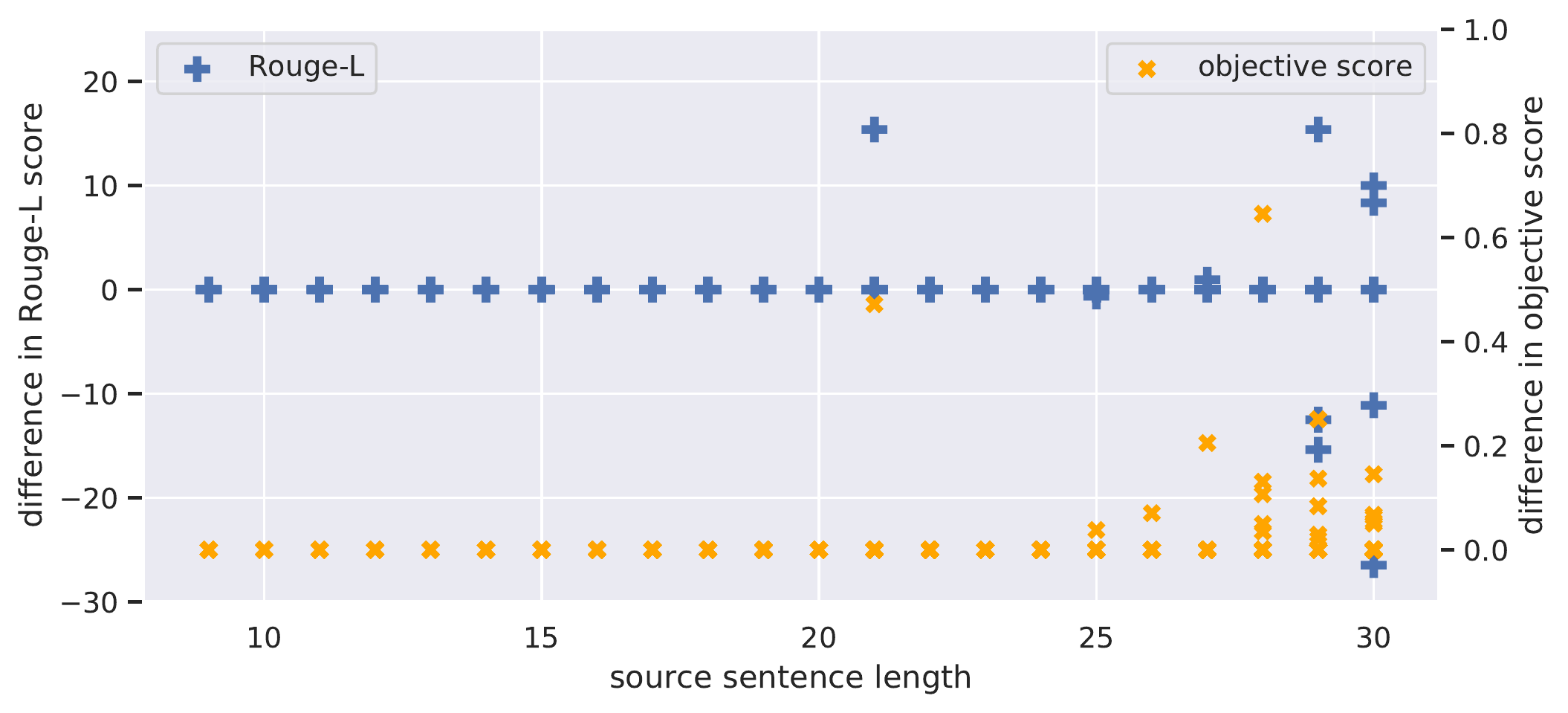}
    \caption{Orange crosses show the objective score optimized by exhaustive search minus the objective score optimized by FCHC. Blue pluses show the \Rouge-L difference between exhaustive search and FCHC. Plotted for the 1135 instances in the headline generation test set, where the source sentence has 30 words or fewer.}
    \label{fig:exhaustive}
\end{figure}

\textbf{Search Algorithm.} In Figure~\ref{fig:exhaustive}, we compare our FCHC with the theoretical optimum on short sentences where exhaustive search is tractable. For only 3\% of the instances with source sentence length between 25 and 30 words, our FCHC algorithm does not find the global optimum. In 21\% of those cases, the better objective score leads to a higher \Rouge-L score. This shows that FCHC with restarts is a powerful enough search algorithm for word extraction-based sentence summarization.

\textbf{Positional Bias.}
\begin{figure}[t]
    \centering
    \vspace{-.2cm}
    \includegraphics[width=0.49\textwidth]{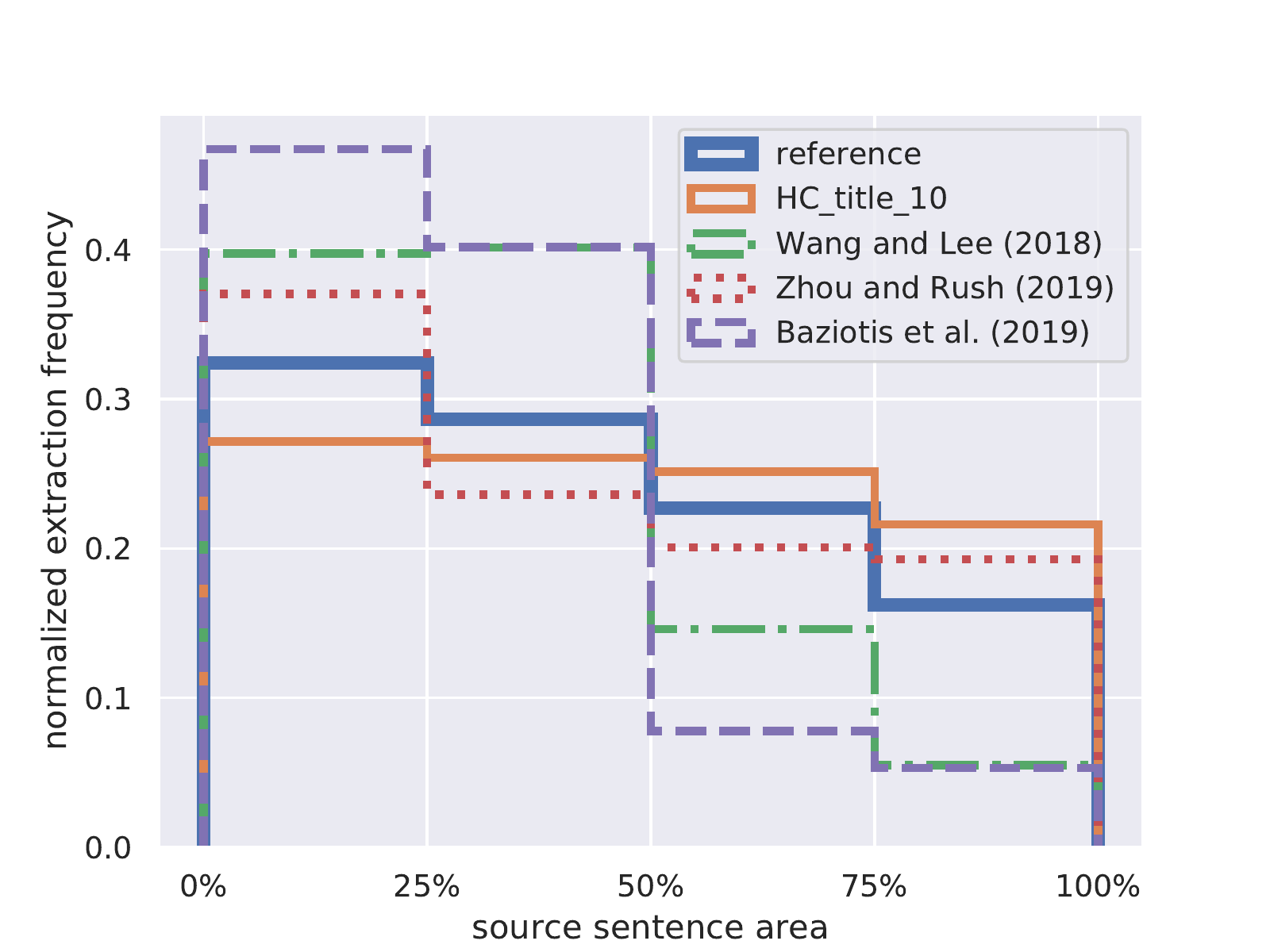}
    \caption{Positional bias for different systems, calculated for the headline generation test set. The source sentence is divided into four areas: 0--25\%, 25--50\%, 50--75\%, and 75-100\% of the sentence. The $y$-axis shows the normalized frequency of how often a word in the summary is extracted from one of the four source sentence areas.}
    \label{fig:position}
\end{figure}
We analyze the positional bias of each algorithm by plotting the normalized frequency of extracted words within four different areas of the source sentence. As shown in Figure \ref{fig:position}, the extraction positions of words in the reference headlines are slightly skewed towards the beginning of the source sentence. Our hill-climbing algorithm performs distributed edits over the sentence, which is reflected in the flatter graph across the source sentence areas. By contrast, beam search \cite{zhou2019simple} is more biased towards the first quarter of the source sentence. Cycle consistency models \cite{wang-lee-2018-learning, baziotis2019seq} show a strong bias towards the first half of the source sentence. We suspect that the reconstruction decoder is easily satisfied with the beginning of the source sentence as the discrete latent variable, because of its autoregressive decoding.

\textbf{Case Study.}
We show example summaries generated by our system in Figure \ref{fig:examples}. We see that the \textit{HC\_title} models indeed learn the style of headlines, known as \textit{headlinese}. As shown, \textit{HC\_title} often uses simple tense and drops articles (e.g., ``a'' and ``the''). The summaries generated by \textit{HC\_article} tend to waste word slots by including an uninformative determiner. 
 
It is also seen that we can control the length in an explicit way. Comparing \textit{HC\_title} with desired lengths of 8 and 10, we see that the additional two words are used to include more information, such as the day of the meeting in Example~2 or the gender of the injured person in Example~3.

\begin{table}[!t]
  \footnotesize
  \begin{tabular}{|p{\linewidth}|}   
 \midrule
\textbf{1. Input:} a german registered container ship ran aground at the entrance to the french port of le havre early tuesday , but authorities said there were no casualties . \\
\textbf{Reference}: container ship runs aground in french port \\
\textbf{\textit{HC\_article\_10}}: a container ship ran aground but there were no casualties \\
\textbf{\textit{HC\_title\_10}}: container ship ran aground at french port but no casualties \\
\textbf{\textit{HC\_title\_8}}: ship ran aground at french port no casualties \\
 \midrule

\textbf{2. Input:} fidel castro , cuba's president of the council of state , met with a chinese delegation here tuesday . \\
\textbf{Reference:} castro meets chinese official \\
\textbf{\textit{HC\_article\_10}}: fidel castro cuba 's president met with a chinese delegation \\
\textbf{\textit{HC\_title\_10}}: fidel castro cuba 's president met with chinese delegation tuesday \\
\textbf{\textit{HC\_title\_8}}: fidel castro 's president met with chinese delegation \\
 \midrule

\textbf{3. Input:} two grenades exploded near a national police station monday , slightly injuring one woman , news reports said . \\
\textbf{Reference:} two grenades explode near spanish police station \\
\textbf{\textit{HC\_article\_10}}: two grenades exploded near a police station injuring one woman \\
\textbf{\textit{HC\_title\_10}}: two grenades exploded near a police station injuring one woman \\
\textbf{\textit{HC\_title\_8}}: two grenades exploded near police station injuring one \\
 \midrule

\end{tabular}
\vspace*{-0.3cm}
  \caption{Example summaries for headline generation test set.}
  \label{fig:examples}
\end{table}
\section{Conclusion}
We proposed a novel word-extraction model for sentence summarization that generates summaries by optimizing an objective function of language fluency and semantic similarity. A hard length constraint is also imposed in our objective function. In a controlled experiment, our model achieves better performance than strong baselines on headline generation and DUC2004 datasets.

\section*{Acknowledgments}
We acknowledge the support of the Natural Sciences and Engineering Research Council of Canada (NSERC), under grant Nos.~RGPIN-2019-04897, and RGPIN-2020-04465. 
Lili Mou is also supported by AltaML, the Amii
Fellow Program, and the Canadian CIFAR AI Chair Program. This research was enabled in part by the support of Compute Canada (\url{www.computecanada.ca}). 

\bibliography{acl2020}
\bibliographystyle{acl_natbib}

\end{document}